\pdfoutput=1

\documentclass[preprint,12pt,times]{elsarticle}

\usepackage[left=1.6cm,right=1.6cm,top=2.54cm,bottom=2.54cm]{geometry}
\usepackage{graphicx}
\usepackage{booktabs}
\usepackage{siunitx}
\usepackage{amsmath,amssymb}
\usepackage{microtype}
\usepackage{textcomp}
\usepackage{placeins}
\usepackage{xurl}
\usepackage{wrapfig}
\makeatletter
\def\ps@pprintTitle{}
\makeatother

\journal{}

\begin{document}

\begin{frontmatter}
\title{Developing and Validating a High-Throughput Robotic System for the Accelerated Development of Porous Membranes}

\author[1]{Hongchen Wang}
\author[2]{Sima Zeinali Danalou}
\author[2]{Jiahao Zhu}
\author[2]{Kenneth Sulimro}
\author[3]{Chaewon Lim}
\author[4]{Smita Basak}
\author[1]{Aim\'ee Tai}
\author[4]{Usan Siriwardana}
\author[1]{Jason Hattrick-Simpers\corref{cor1}}
\author[2]{Jay Werber\corref{cor1}}

\affiliation[1]{organization={Department of Materials Science and Engineering, University of Toronto},
            addressline={27 King's College Cir}, 
            city={Toronto},
            postcode={M5S 1A1}, 
            state={ON},
            country={Canada}}
\affiliation[2]{organization={Department of Chemical Engineering, University of Toronto},
            addressline={27 King's College Cir}, 
            city={Toronto},
            postcode={M5S 1A1}, 
            state={ON},
            country={Canada}}
\affiliation[3]{organization={Department of Engineering Science, University of Toronto},
            addressline={27 King's College Cir}, 
            city={Toronto},
            postcode={M5S 1A1}, 
            state={ON},
            country={Canada}}
\affiliation[4]{organization={Department of Mechanical Engineering, University of Waterloo},
            addressline={200 University Ave W}, 
            city={Waterloo},
            postcode={N2L 3G1}, 
            state={ON},
            country={Canada}}

\cortext[cor1]{jason.hattrick-simpers@utoronto.ca; jay.werber@utoronto.ca}

\begin{abstract}
\begin{wrapfigure}{r}{0.55\textwidth}
  \vspace{-2em}
  \includegraphics[width=\linewidth]{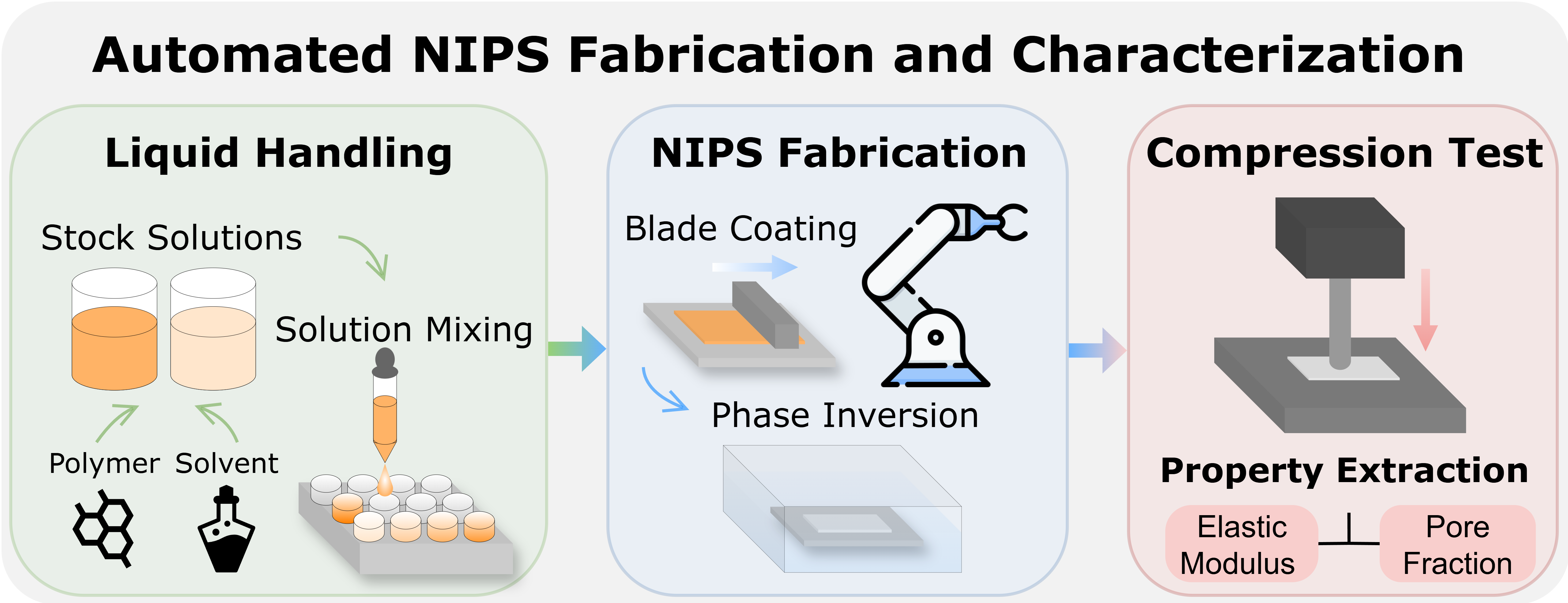}
\end{wrapfigure}
The development of porous polymeric membranes remains a labor-intensive process, often requiring extensive trial and error to identify optimal fabrication parameters. In this study, we present a fully automated platform for membrane fabrication and characterization via nonsolvent-induced phase separation (NIPS). The system integrates automated solution preparation, blade casting, controlled immersion, and compression testing, allowing precise control over fabrication parameters such as polymer concentration and ambient humidity. The modular design allows parallel processing and reproducible handling of samples, reducing experimental time and increasing consistency. Compression testing is introduced as a sensitive mechanical characterization method for estimating membrane stiffness and as a proxy to infer porosity and intra-sample uniformity through automated analysis of stress-strain curves. As a proof of concept to demonstrate the effectiveness of the system, NIPS was carried out with polysulfone, the green solvent PolarClean\textsuperscript{\textregistered}, and water as the polymer, solvent, and nonsolvent, respectively. Experiments conducted with the automated system reproduced expected effects of polymer concentration and ambient humidity on membrane properties, namely increased stiffness and uniformity with increasing polymer concentration and humidity variations in pore morphology and mechanical response. The developed automated platform supports high-throughput experimentation and is well-suited for integration into self-driving laboratory workflows, offering a scalable and reproducible foundation for data-driven optimization of porous polymeric membranes through NIPS.
\end{abstract}


\end{frontmatter}

\twocolumn
\section{Introduction}
\label{sec1}
Porous polymeric membranes are widely used in separation processes, including ultrafiltration (UF), gas separation, and as support layers for thin-film composite (TFC) membranes used in nanofiltration (NF) and reverse-osmosis (RO) \cite{WARSINGER2018209,Park2017, C9RA07114H}. Emerging high-pressure reverse osmosis (HPRO) processes extend the operation to pressures of 120+~bar, requiring support membranes that can withstand substantial compressive stresses \cite{Davenport2018}. As a result, developing porous membranes with robust and tunable mechanical properties is essential to ensure reliable performance in pressure-driven separation processes \cite{Davenport2018}. 

Currently, most commercial porous membranes are manufactured using nonsolvent-induced phase separation (NIPS) due to its processing simplicity, scalability, and cost-effectiveness \cite{Tan2019, LIU20111}. In the NIPS process, polymer solution is cast onto a substrate and immersed in a nonsolvent bath to initiate the solvent-nonsolvent exchange and subsequent phase separation to form a solid membrane \cite{LIU20111, Mulder1996}. Despite its advantages, the process is highly sensitive to changes in composition, environmental conditions, and processing parameters, often resulting in substantial variability in membrane structure and performance \cite{Guillen2011, Jung2016, Zheng2006}. 

The kinetics of this demixing process strongly influence the resulting membrane structure \cite{Guillen2011}. Specifically, rapid demixing typically forms finger-like macrovoids, which offer high flux but poor mechanical performance, while delayed demixing forms sponge-like structures, which are mechanically robust but less permeable \cite{Guillen2011}. Key fabrication parameters, including polymer concentration, solvent and nonsolvent types, temperature, humidity, and casting thickness, play critical roles in governing membrane morphology and performance \cite{Guillen2011, MOSQUEDAJIMENEZ2004209, STRATHMANN1975179}.

While many studies have examined how these fabrication parameters influence membrane structure and performance \cite{Jung2016, Zheng2006, Zheng20062, Dong2021,ALEXANDRE20001}, optimizing these variables through traditional experimentation remains slow and labor-intensive. Manual workflows are often inconsistent and difficult to reproduce, particularly given the complex and sensitive dynamics of NIPS \cite{Guillen2011, Garcia2020}. Furthermore, membrane characterization often involves multiple time-consuming steps, such as filtration testing and scanning electron microscopy (SEM) \cite{Hobaib2015, Abdullah2014}. These limitations present a significant barrier to rapid iteration and systematic exploration of the NIPS fabrication space.

To overcome challenges similar to those faced in NIPS, high-throughput experimentation (HTE) and self-driving laboratories (SDLs) have emerged as promising paradigms for accelerated materials development \cite{Potyrailo2011, STACH20212702}. HTE enables rapid and systematic exploration of large combinatorial design spaces by automating the fabrication and/or characterization of samples under a wide range of conditions \cite{Potyrailo2011}. While we are not aware of a high-throughput system for NIPS, notable innovations in HTE for membrane science have been demonstrated in areas such as diafiltration, dead-end filtration, and membrane electrode assembly \cite{OUIMET2022119743, VANDEZANDE2005305, D2CY00873D}. 

SDLs take automated experimentation one step further by integrating HTE with real-time data analysis and algorithmic decision-making to iteratively synthesize, test, and optimize material systems with minimal human intervention \cite{STACH20212702, Tom2024, Abolhasani2023}. In other words, SDLs close the loop during automated materials research and development, which can dramatically accelerate materials development and optimization for design spaces with many dimensions where brute-force HTE would be onerous and inefficient. For example, Snapp et al. \cite{Snapp2024} demonstrated an SDL framework that autonomously performed over 25,000 compression tests on 3D-printed polymer structures, efficiently exploring an 11-dimensional design space via robotic experimentation and Bayesian optimization to identify structures with exceptional energy absorption properties. This approach highlights how integrating closed-loop automation with compression testing can yield rapid, data-rich insights into structure-property relationships, even in complex, high-dimensional materials systems. Interest in SDLs has increased markedly in recent years, with applications now developed in pharmaceuticals, catalysis, and materials discovery \cite{Mennen2019, MA2025, Szymanski2023}. 

This manuscript presents our work towards a NIPS-based SDL for the accelerated development of porous membranes. In particular, we have developed an automated NIPS fabrication and testing platform (i.e., an HTE system for NIPS) that integrates liquid handling, robotic blade casting, controlled environmental conditions, and mechanical testing into a single automated system. By combining automated sample preparation with rapid mechanical compression testing, the system allows systematic exploration of the membrane design space under controlled and reproducible conditions. A key innovation of this platform is the use of compression testing as a high-throughput proxy for membrane performance. Drawing from previous research on porous mechanical systems \cite{Gongora2020, Costa2008, Feng2019}, we interpret the stress-strain response of membrane samples to extract two performance-relevant metrics: 1) elastic modulus as a measure of stiffness and 2) normalized plateau length as a proxy for porosity. As a proof of concept exploration using the automated NIPS fabrication platform, we investigate how polymer concentration and ambient humidity influence membrane morphology and mechanical performance, using polysulfone (PSf), the green solvent PolarClean\textsuperscript{\textregistered}, and water as the polymer, solvent, and nonsolvent, respectively. Multiple tests across each membrane demonstrate intra-sample consistency, with comparisons to manually fabricated membranes used to validate the membrane uniformity achieved through automation. The presented system is now capable of high-throughput synthesis and characterization of porous membranes through NIPS, achieving a throughput of less than an hour per sample from formulation to mechanical testing. Our future work will "close the loop" by integrating active learning algorithms and orchestrations to enable full SDL workflows. 

\section{Materials and Methods}
\label{sec2}

\subsection{Materials}
The chemicals used in this study are polysulfone (PSf) (Sigma-Aldrich, $M_w$ - 35000) and PolarClean\textsuperscript{\textregistered} (Rhodiasolv PolarClean\textsuperscript{\textregistered}). The commercially available tools used to build the automated platform are a liquid handling robot (Opentrons OT-2), a robotic arm (Ufactory xArm7), and a compression tester (TestResources 140). 

\subsection{Membrane Fabrication}
Membranes were fabricated from PSf-PolarClean\textsuperscript{\textregistered} solutions. Initially, stock solutions were manually prepared by dissolving PSf pellets in PolarClean\textsuperscript{\textregistered} at 80~ºC with continuous stirring overnight, ensuring complete dissolution. The prepared polymer solution was sonicated for 30~min and allowed to rest for several hours to eliminate air bubbles. The mixed polymer solution was then dispensed into a uniform line onto a clean stainless steel coupon. A doctor blade was slid across the coupon to cast the polymer solution into a uniform film at a controlled speed of 1~cm/s. Humidity control was managed by introducing laminar flow of dry nitrogen from a nitrogen blower positioned above the freshly cast films. If further analyses were needed, the fabricated and tested membranes were immersed in fresh deionized (DI) water and stored in the refrigerator to prevent bacterial growth. This procedure shows the general workflow used in both manual and automated fabrication. Detailed descriptions of the robotic automation are provided in Section \ref{Design of Automated NIPS System}.

\subsection{Scanning Electron Microscopy Characterization}
Scanning electron microscopy (SEM) was used to evaluate the microstructure of membranes fabricated under different experimental conditions. Both cross-sectional and top-view images were obtained to assess internal pore morphology and surface features. All SEM imaging was performed using a Hitachi SU7000 scanning electron microscope.

Membrane samples were prepared by the freeze-fracturing method to preserve their internal morphology and minimize deformation artifacts. Before fracturing, the membranes were subjected to a sequential solvent exchange treatment to ensure pore preservation: samples were immersed in DI water for 10~min, then transferred to 25\% isopropyl alcohol (IPA) for 10~min, pure IPA for 5~min, and finally hexane for 10~min. The sample was then firmly clamped and immersed in liquid nitrogen for 1~min to 2~min and fractured to obtain a clean cross-sectional surface. 

The fractured samples were subsequently trimmed to suitable dimensions and mounted on SEM stubs using conductive carbon tape. The corners were dabbed with carbon paint to minimize the charging effect during imaging. A thin platinum layer (1.8~nm to 3~nm thick) was deposited on the exposed cross-sectional or top surfaces using a Leica ACE600 system, preventing charging and enhancing image clarity during SEM imaging. Imaging was carried out at an accelerating voltage of 1~kV to 1.5~kV, using appropriate magnifications to clearly visualize the thickness, pore morphology, and structural characteristics of the membranes. 

\subsection{Pore Size Distribution Analysis}
Pore segmentation was performed using Fiji/ImageJ (v1.54f) and the Trainable Weka Segmentation plugin, followed by watershed separation. Image enhancement and binarization followed a previously published protocol developed by our group \cite{SIMA2025}. To correct for pore narrowing caused by sputter coating and estimate the uncoated pore size, a 1.8~nm dilation (corresponding to the applied coating thickness on the top surfaces) was applied to the binary pore masks, as described in our previous work \cite{SIMA2025}. Pore size distributions (PSD) were calculated from three replicate SEM images per sample. Area-weighted histograms were generated by assigning each pore a circular area, binning the diameters at 0.5~nm intervals, and normalizing by the total image area. The mean distribution across replicates was reported with the corresponding standard error.

\section{Results} 

\subsection{Design of Automated NIPS System}
\label{Design of Automated NIPS System}
\begin{figure*}[ht]
\centering
\includegraphics[width=\linewidth]{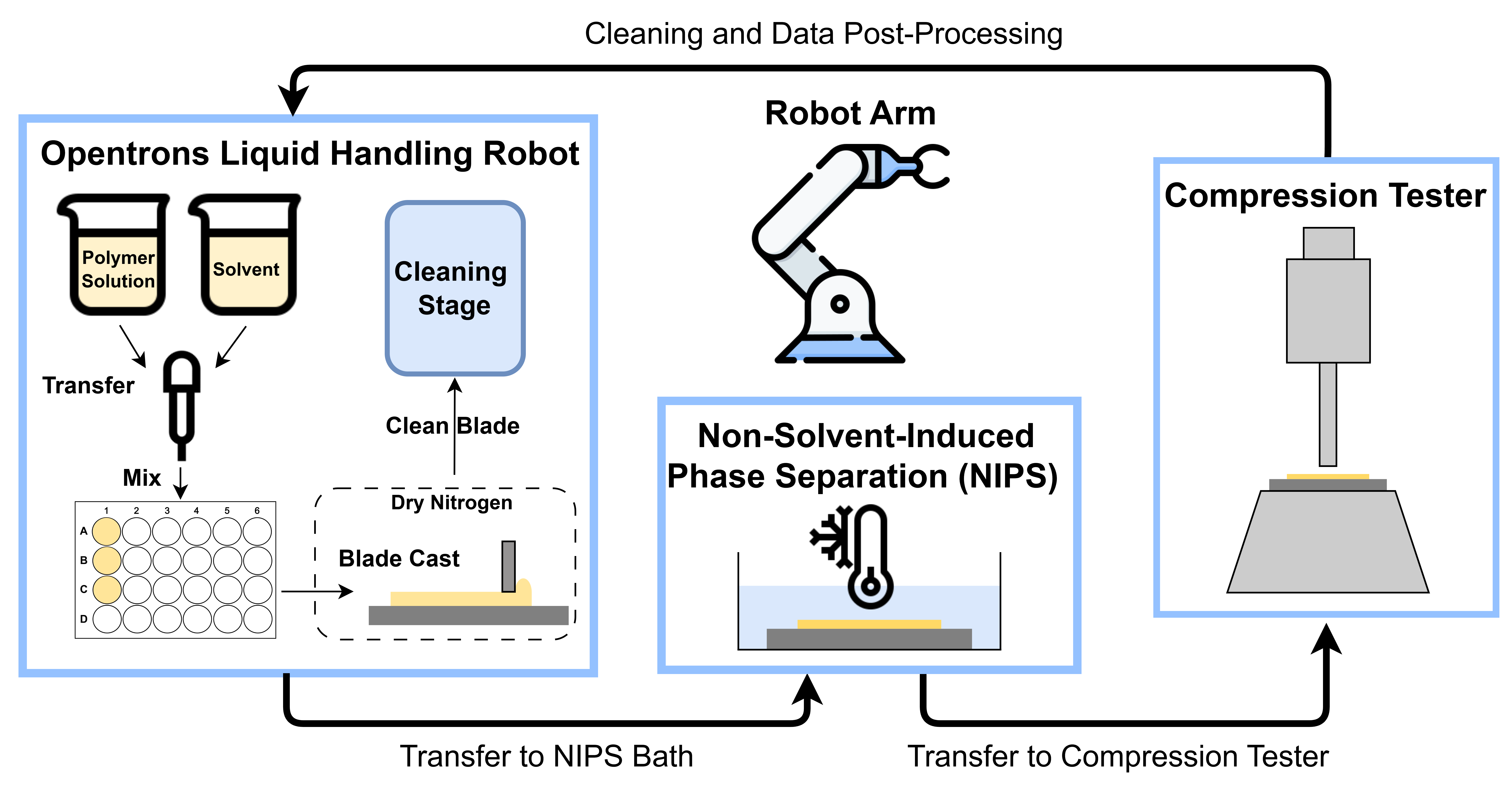}
\caption{Planned workflow for a NIPS-based self-driving lab (SDL). This workflow demonstrates an iterative process between a liquid handling robot, a temperature-controlled coagulation bath, and a compression tester, interconnected by a robotic arm. The liquid handling robot transfers and mixes the precursor solutions. The robotic arm performs blade casting and sample transferring between the different modules. NIPS takes place in the coagulation bath. Once a membrane is fabricated, the compression tester performs a series of tests on the sample to characterize its intra-sample consistency and mechanical properties. Upon post-processing and analysis, the next set of experiments can be initiated and operated iteratively.}
\label{HTE_workflow.png}
\end{figure*}

Figure~\ref{HTE_workflow.png} provides an overview of the automated NIPS platform. The integrated robotic system consists of three essential modules: a liquid handling robot for polymer solution preparation, a temperature-controlled coagulation bath for NIPS, and a compression tester for mechanical characterization. These modules are interconnected by a robotic arm, which performs blade casting and transfers samples, minimizing human intervention and ensuring reproducibility and consistency. Initially, stock solutions (i.e., polymer solutions, solvents) are manually prepared and loaded into the liquid handling robot. The robot is programmed to mix polymer solutions of targeted compositions, after which the robotic arm casts the prepared solution into a film using a doctor blade. The cast film is then immersed in the coagulation bath, where phase inversion occurs. Following NIPS, the robotic arm transfers the membrane to the compression tester to obtain the stress-strain curves. Mechanical properties such as stiffness and pore fraction are extracted using our designed data-processing pipeline. The entire robotic system operates within a ductless fume hood, minimizing exposure risks and ensuring operational safety. All equipment used for the system is listed in the Supporting Information.

\begin{figure*}[!ht]
\centering
\includegraphics[width=\linewidth]{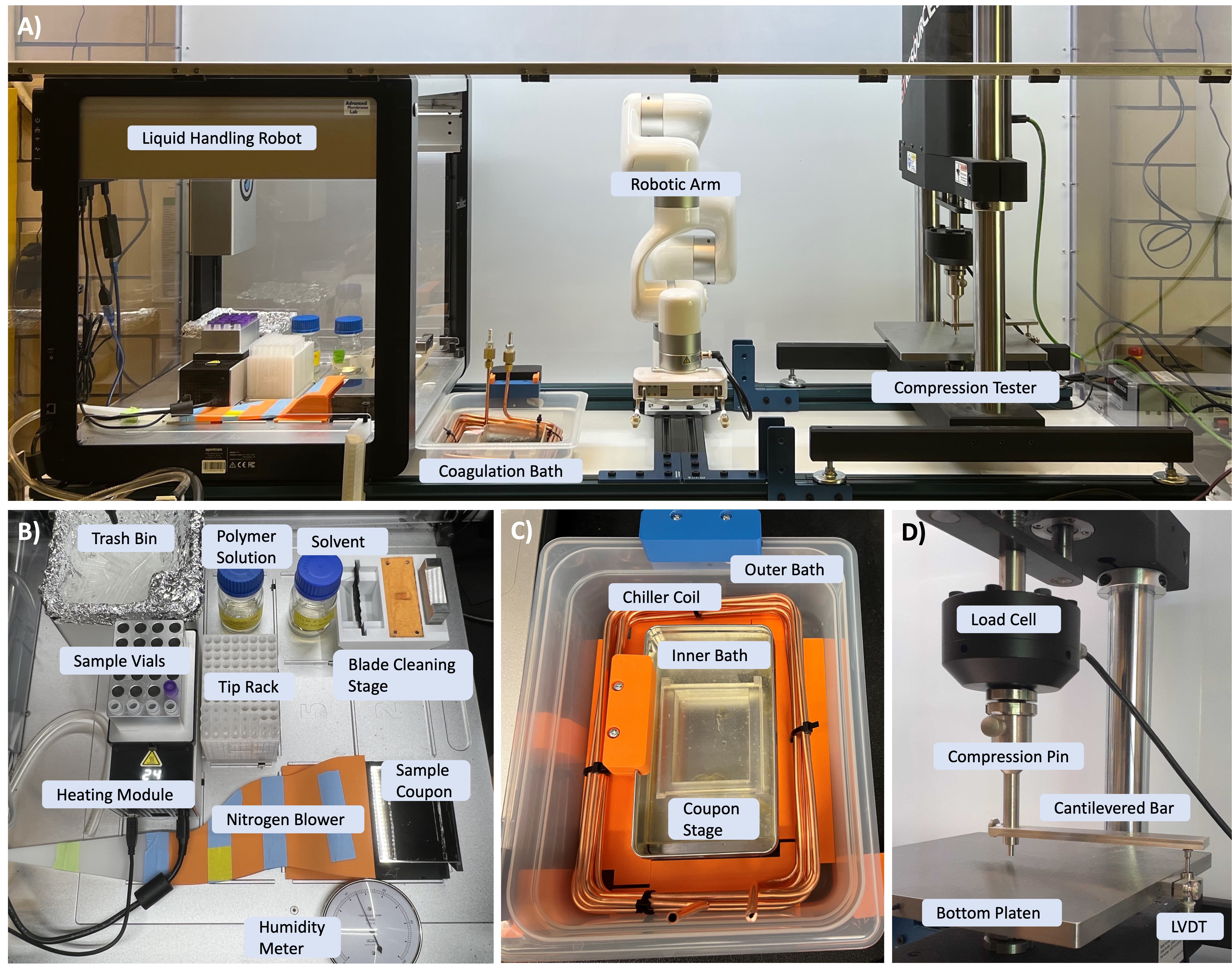}
\caption{Experimental setup used for automated membrane fabrication and mechanical testing. (A) Overall system in a ductless fume hood. (B) The liquid handling robot is configured for polymer solution preparation. It uses pipette tips to transfer and mix the precursor solutions into the vials. The well plate can be heated to achieve more homogeneous mixing. The nitrogen blower introduces a laminar dry nitrogen flow over the cast film to reduce the humidity. The blade cleaning stage holds and cleans the doctor blade after each blade casting. (C) The temperature-controlled coagulation bath uses a copper coil that is cooled by a circulating chiller to achieve temperature control during phase separation. (D) The compression tester uses a 5 mm diameter flat punch to characterize the mechanical properties of the fabricated membranes. It is equipped with a load cell for force measurement and an external linear variable differential transformer (LVDT) for precise small displacement measurement.}
\label{Module_pics.jpg}
\end{figure*}

Detailed descriptions of each experimental module are provided in Figure~\ref{Module_pics.jpg}. Figure~\ref{Module_pics.jpg}A shows the overall layout of the system inside a ductless fume hood. Figure~\ref{Module_pics.jpg}B shows the layout of the liquid handling robot. Upon loading the stock solutions and solvents, the liquid handling robot is programmed to prepare polymer solutions at precisely controlled concentrations ranging from 10 to 17~wt\%, with polymer concentration serving as a critical controllable parameter that impacts membrane morphology and mechanical properties. The system incorporates heated sample vials to maintain the homogeneity of the polymer solution. The polymer solution is dispensed onto a sample coupon by the liquid handling robot, after which the robotic arm picks up the doctor blade and performs a speed-controlled blade casting operation to produce uniform polymer films. The nitrogen blower positioned next to the casting station delivers laminar dry nitrogen flow, reducing the ambient humidity and thereby controlling the pre-immersion conditions. Although only dry nitrogen was used in this study, the relative humidity of this inlet gas stream can be readily tuned by mixing dry and wet gases to adjust the humidity of the film environment. A dedicated blade cleaning stage ensures that residual polymer solution on the doctor blade is removed between each casting step, minimizing cross-contamination and enhancing reproducibility. 

Figure~\ref{Module_pics.jpg}C shows the temperature-controlled coagulation bath. The bath is designed with two compartments: 1) an outer bath containing a copper cooling coil connected to a circulating chiller and 2) an inner bath where phase separation occurs. Temperature control of the coagulation bath enables systematic exploration of the effect of temperature on the morphology and performance of the membranes. In this study, the bath temperature was held constant; future work will incorporate temperature variation along with optimization of other processing parameters. After blade casting, the robotic arm transfers the sample into the inner bath, where phase inversion occurs. For the samples in this study, membrane formation occurs for approximately 30~min in the bath. In future work, two baths will be used for initial coagulation and full solvent removal, respectively. The automated NIPS workflow can be viewed in the video in the Supporting Information.

Figure~\ref{Module_pics.jpg}D shows the setup of the compression tester module. To accurately assess the mechanical properties of membrane samples, the tester uses a 5 mm diameter flat punch connected to a load cell to measure the applied force. Due to the thinness (50~µm to 120~µm) and mechanical sensitivity of the membranes, displacement is measured using an external linear variable differential transformer (LVDT). The LVDT offers a high-resolution measurement range of ±1270~µm with a linearity of ±3~µm to ensure detection of subtle differences in the mechanical responses of the membranes. After placing the sample on the bottom platen, a triggering signal automatically sent from the computer initiates the test, generating a detailed force-displacement dataset, which is later converted to stress-strain (stress = force / compression pin area; strain = displacement / thickness). This data is automatically processed using our in-house noise reduction and feature extraction algorithm, which will be discussed in Section \ref{Compression Testing for Mechanical Proxy Characterization}.

\subsection{Compression Testing for Mechanical Proxy Characterization}
\label{Compression Testing for Mechanical Proxy Characterization}

\begin{figure*}[!ht]
\centering
\includegraphics[width=\linewidth]{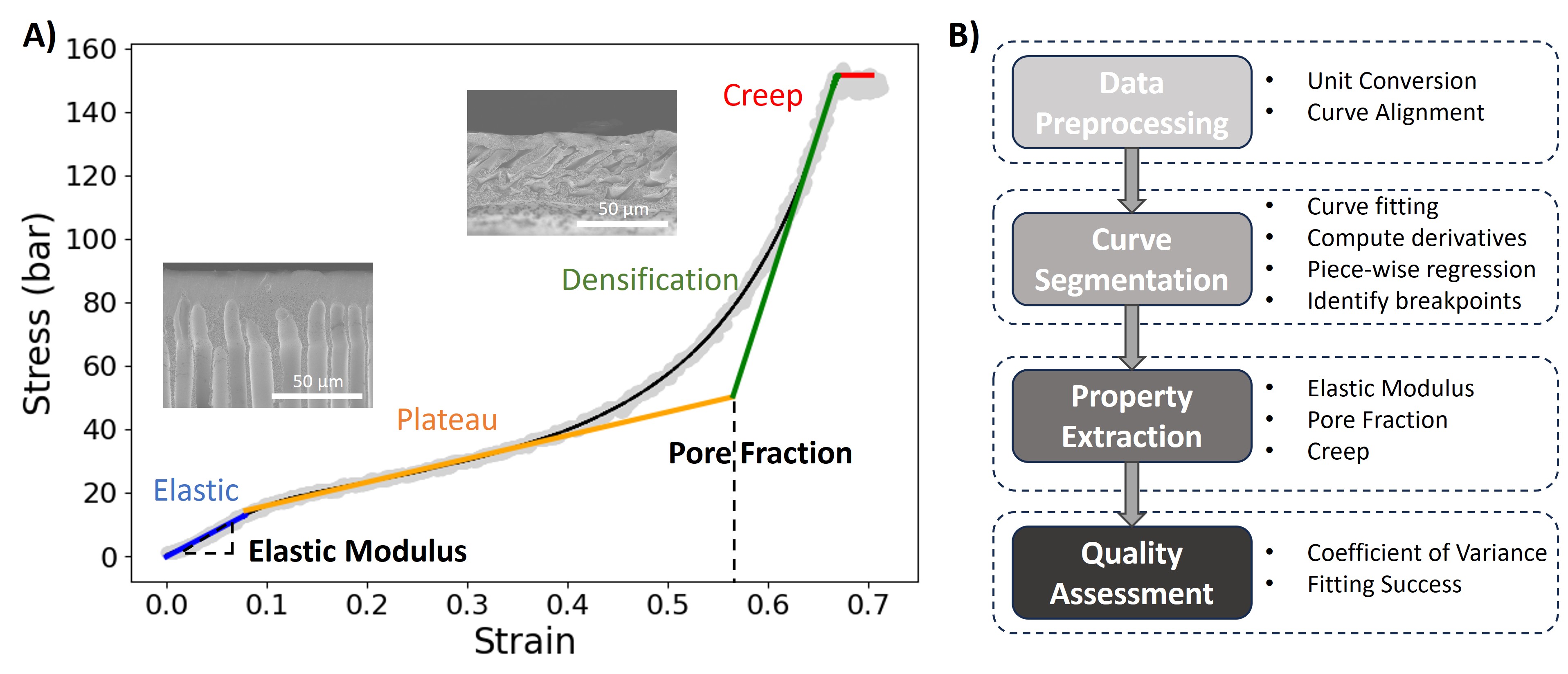}
\caption{Demonstration of the automated stress-strain curve analysis. (A) Automatic segmentation of a stress-strain curve (engineering stress and engineering strain) using a designed pipeline in Python to identify key mechanical regions: elastic (blue), plateau (orange), densification (green), and creep (red). This segmentation enables estimation of the elastic modulus and pore fraction of the membrane, which for this sample are 166.1~bar and 0.57, respectively. Inset SEM images show the microstructure of a membrane before and after compression. (B) Data-processing pipeline for mechanical property extraction, consisting of four key stages: (1) Data Preprocessing, including unit conversion and curve alignment; (2) Curve Segmentation, where breakpoints are computed using derivatives and piecewise regression; (3) Property Extraction, determining elastic modulus, yield strength, densification, and creep properties; and (4) Quality Assessment, evaluating the quality and intra-sample consistency through statistical metrics such as the coefficient of variance (CV) and fitting success criteria.}
\label{Data_processing.jpg}
\end{figure*}

Compression testing was used as a proxy measurement for rapid assessment of membrane homogeneity and mechanical properties. While traditionally used in the study of foams and porous ceramics \cite{Thiyagarajan2022, Chen2017}, this approach has also been applied to polymeric membranes to infer porosity and mechanical integrity from the bulk mechanical response \cite{AGHAJANI2017293, Aghajani2018, Gibson_2003}. This method efficiently captures subtle variations in membrane structure by analyzing their mechanical responses to compressive loading. As illustrated in Figure~\ref{Data_processing.jpg}A, a typical stress-strain curve of a porous material exhibits three distinct regions: 1) elastic, 2) plateau, and 3) densification, correlated with specific mechanical behaviors and microstructural deformations at different stages under compressive loading \cite{Costa2008, Thiyagarajan2022, AGHAJANI2017293, Gibson_2003}. A creep region is also observed if a creep test (constant force over time) is performed. Initially, the elastic region characterizes the effective stiffness of the membrane, quantifying its resistance to deformation under compressive loading. After the elastic limit, the membrane enters the plateau region, where a lower slope is observed due to progressive pore deformation and collapse. Once the pores are fully collapsed, the curve enters the densification region, where the stress increases sharply as the solid matrix and substrate dominate the load-bearing behavior. Pore fraction, in particular, is identified as the strain at which the linear fits to the plateau and densification regions intersect. Throughout the rest of this study, we use pore fractions estimated by compression testing as a proxy of membrane porosity, as opposed to bulk porosity measurements through traditional gravimetric approaches. Previous studies have demonstrated the validity of this approach for membranes and other porous materials \cite{Aghajani2018, Miralbes2023, Xu2024}. SEM images inset in Figure~\ref{Data_processing.jpg}A visually show the morphological changes in the membrane before and after compression. To characterize the intra-sample homogeneity and mechanical consistency, multiple compression tests are performed at different positions across the membrane sample area.

Figure~\ref{Data_processing.jpg}B details the automated data-processing pipeline developed to systematically analyze the compression data. The pipeline involves four key stages: (1) data preprocessing, including unit conversion and curve alignment; (2) curve segmentation using spline fitting, derivative analysis, and piece-wise regression to identify distinct mechanical regions; (3) property extraction, calculating key properties such as elastic modulus, yield strength, pore fraction, and creep strain from segmented curves; and (4) quality assessment, evaluating the intra-sample consistency through statistical metrics like coefficient of variance (CV) and fitting success criteria. These extracted mechanical parameters subsequently serve as key inputs for selecting the next set of experimental parameters. The automated curve segmentation and statistical evaluation allow systematic assessment of membrane stiffness, pore fraction, and intra-sample uniformity, providing actionable feedback for iterative optimization in SDL workflows.

\subsection{Validation of Automated NIPS System}
The automated NIPS platform improves experimental throughput by streamlining solution preparation and parallelizing key fabrication steps. In manual workflows, each polymer solution must be prepared separately by weighing out polymers and solvents, followed by extended stirring, heating, and degassing, often requiring overnight processing for each batch. In contrast, the automated system uses an automated dilution strategy, where a concentrated polymer stock solution is prepared once and subsequently diluted by the liquid handling robot into lower concentrations via precise volumetric control while avoiding air bubble formation during mixing. This dilution-based approach eliminates the need to manually prepare and process each concentration individually, enabling rapid and reproducible preparation of different compositions. While the robot prepares new formulations, the robotic arm can simultaneously perform blade casting and sample transfer, enabling different operations to run in parallel to further increase throughput. The entire process from solution preparation to membrane mechanical testing takes less than an hour per sample, which can be further decreased by preparing multiple samples in parallel.

\begin{figure}[!ht]
\centering
\includegraphics[width=\linewidth]{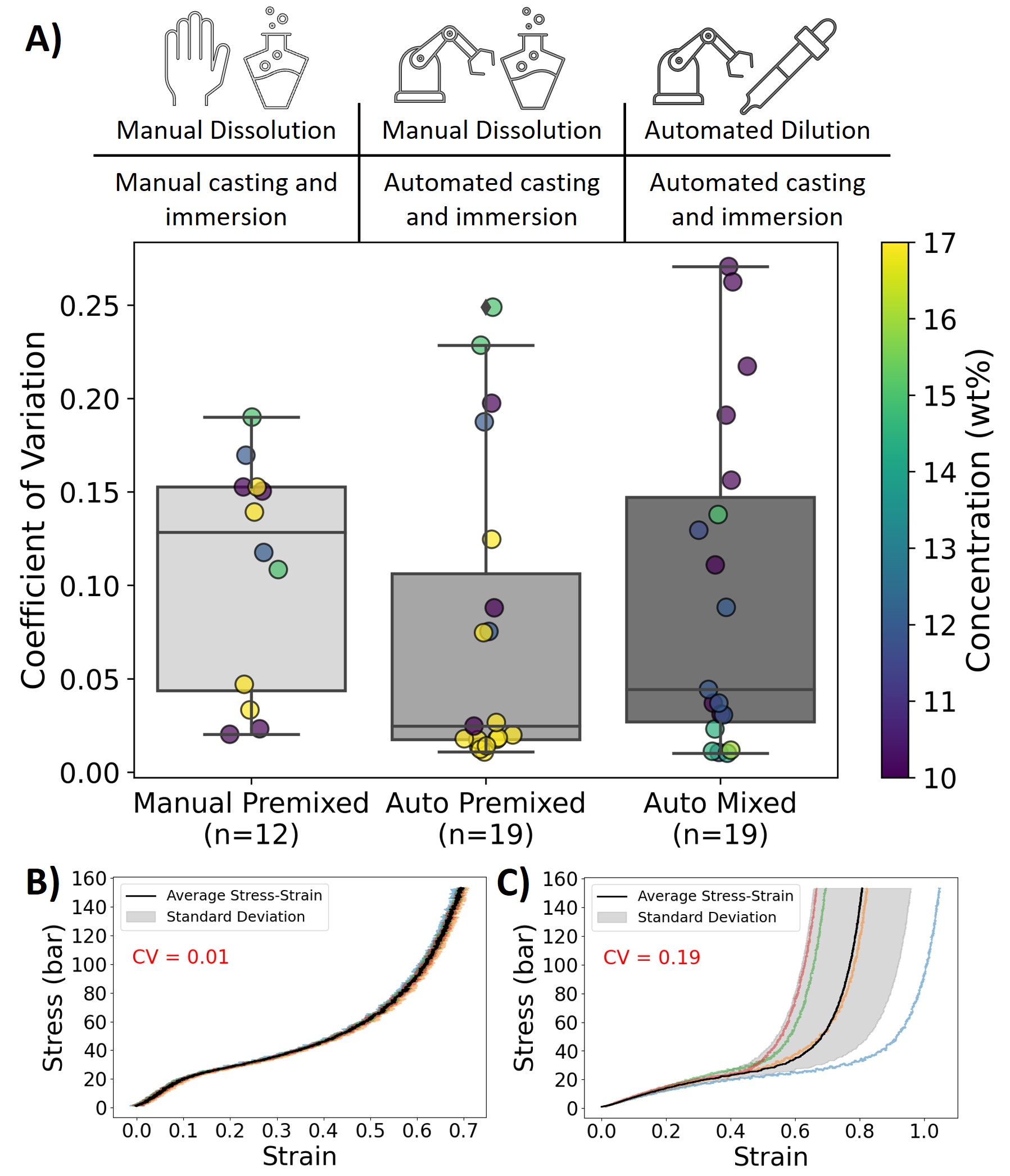}
\caption{Intra-sample consistency analysis using the coefficient of variance (CV). A) Box plot showing the CV of stress-strain curves across three fabrication methods: 1) Manual Premixed (manual solution preparation and fabrication), 2) Auto Premixed (manual solution preparation with automated fabrication), and 3) Auto Mixed (fully automated mixing and fabrication). The horizontal line inside each box marks the median CV; the box itself spans the inter-quartile range (IQR) (25~\%--75~\%); the “whiskers” extend to 1.5~×~IQR; points beyond the whiskers are outliers. Each dot represents a sample, color-coded by polymer concentration (wt\%). The number of samples (n) for each category is specified in parentheses. Examples of stress-strain curves with B) a low CV obtained from a high polymer concentration sample and C) a high CV obtained from a low polymer concentration sample.}
\label{consistency-with-label.jpg}
\end{figure}

Intra-sample consistency reflects how uniform a membrane's mechanical response is across different regions. For each membrane, multiple compression tests were performed at different locations, and the CV of the curves was calculated as a quantitative measure of intra-sample consistency. Compared to crude measurements such as thickness and porosity and qualitative approaches such as visual comparison of cross-sectional SEM images, CV provides more granular detail of local mechanical heterogeneity. It was calculated by first interpolating all strain-stress curves to a common stress axis, then calculating the average standard deviation of strain across all stress points, normalized by the mean strain. Lower CV values indicate more homogeneous membrane structures and mechanical properties. Figure~\ref{consistency-with-label.jpg} compares the CV values of three preparation methods: 1) Manual Premixed, 2) Auto Premixed, and 3) Auto Mixed. 
    
In the Manual Premixed condition, both solution preparation and casting were performed manually, with each concentration prepared by the conventional dissolution method. In the Auto Premixed condition, the polymer solutions were prepared manually by dissolution but were cast and processed using the automated platform. In Auto Mixed, both solution mixing (via dilution from precursors) and casting were performed entirely by the robotic system. The Auto Mixed condition is, of course, the required fabrication pathway for a NIPS-based SDL, as the fabrication process in this approach is fully automated, except for the manual preparation of the initial stock solutions. Among these groups, the Auto Premixed condition yielded the lowest median CV, indicating high intra-sample consistency when human-prepared solutions were paired with automated fabrication. The Manual Premixed group showed higher variability, indicating inconsistencies of manual casting and immersion steps. The Auto Mixed group exhibited longer whiskers and a broader spread of CV values, particularly at lower polymer concentrations, indicating increased variability in intra-sample consistency under this condition. 

Ultimately, the relative similarities of CV between the three conditions, Manual Premixed, Auto Premixed, and Auto Mixed, validate the ability of the system to reproducibly fabricate porous membranes from a target NIPS condition using a fully automated fabrication pathway. The relatively broad spread in CV for the desired Auto Mixed fabrication pathway is almost entirely confined to the low polymer concentration samples (i.e., 10~wt\%), and is likely due to the challenge of mixing solutions with a large difference in viscosity \cite{Mcdermott2023}. In preparation of solutions with low polymer concentrations, the robot mixes more equal volumes of high-viscosity polymer solution and low-viscosity solvent. This disparity in viscosity can compromise mixing uniformity, leading to local composition gradients or incomplete homogenization \cite{Mcdermott2023}. These inconsistencies can translate into structural and mechanical heterogeneity in the fabricated membrane. Although not implemented in this study, a likely solution for future work is to replace the solvent with a dilute polymer solution (e.g., mix 17~wt\% polymer solution with 10~wt\% polymer solution to target an intermediate concentration) to reduce the difference in viscosity of the precursors.

\subsection{Interpretation from Stress-Strain Behavior}

\begin{figure}[!ht]
\centering
\includegraphics[width=\linewidth]{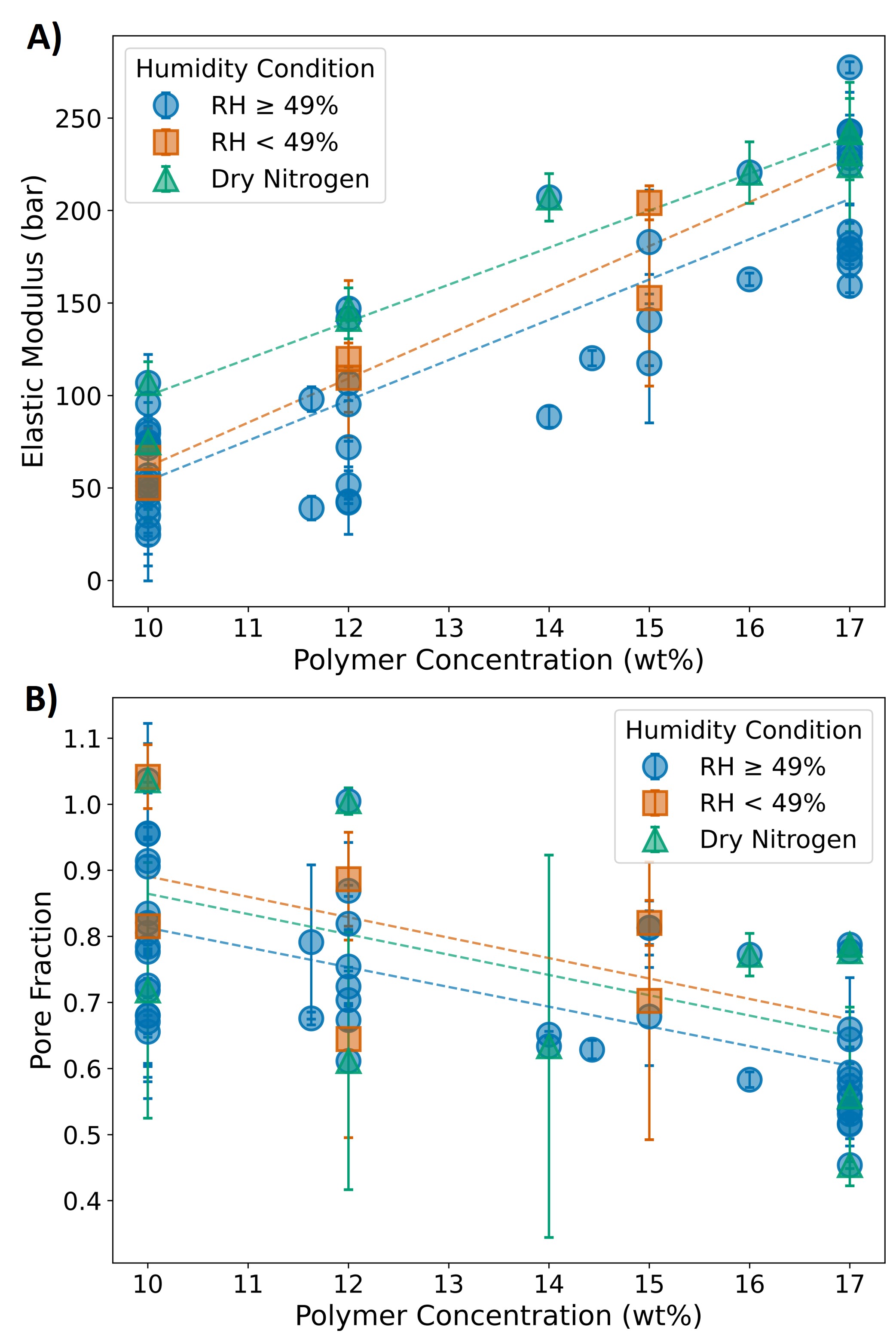}
\caption{Properties extracted from the stress-strain curves using the automated curve segmentation algorithm. (A) Elastic Modulus vs. Polymer Concentration and (B) Pore Fraction vs. Polymer Concentration across different humidity conditions: 1) $RH \geq 49\%$ (blue), 2) $RH < 49\%$ (orange), and 3) with dry nitrogen exposure (green). RH refers to the ambient humidity for experiments that did not use dry nitrogen during NIPS. Lines of best fit are drawn as a guide to the eye.}
\label{modulus_and_pore_fraction_V.jpg}
\end{figure}

The above validation of the system to reproducibly generate porous membranes while varying NIPS conditions (e.g., polymer concentration) suggests that the system is ready for high-throughput experimentation and self-driving lab methodologies. As a proof of concept, we used the system to explore the PSf/PolarClean\textsuperscript{\textregistered}/water NIPS system, centering on compression testing as the main membrane characterization methodology. Figure~\ref{modulus_and_pore_fraction_V.jpg} shows the effect of polymer concentration and ambient humidity on mechanical stiffness and pore fraction of PSf membranes. Both plots show noticeable scatter, likely due to initial experimental variability during the optimization of robotic pipetting parameters and sample handling processes, as well as inherent stochasticity associated with the NIPS process itself. Figure~\ref{modulus_and_pore_fraction_V.jpg}A shows a consistent increase in elastic modulus with higher polymer concentration across all humidity conditions, indicating that higher polymer content leads to denser, mechanically stronger membranes. The membranes cast under dry nitrogen (green) exhibited the highest modulus, while those prepared under high humidity ($RH \geq 49\%$, blue) showed the lowest, reflecting greater porosity and reduced mechanical strength. Intermediate humidity conditions ($RH < 49\%$, orange) resulted in the modulus between these extremes. 

Shown in Figure~\ref{modulus_and_pore_fraction_V.jpg}B, pore fractions estimated from stress-strain curve-fitting tend to decrease with increasing polymer concentration, supporting the formation of denser membranes at higher polymer content. Values above 1 at low concentrations likely result from lateral slippage of the compression pin on soft, low-friction samples, which rotates the cantilever and further compresses the LVDT. Significant scatter and overlap are observed among the different humidity conditions. Unlike the clear humidity-dependent trends observed in elastic modulus, it is difficult to conclude whether there is a correlation between pore fraction and ambient humidity from current data. This suggests that pore formation may be sensitive to subtle localized variations in solvent-nonsolvent exchange kinetics, potentially amplified by minor fluctuations in experimental parameters and processing steps. These observations confirm a key established behavior in membrane fabrication using NIPS: higher polymer concentrations consistently produce stiffer and generally less porous membranes, while humidity can also significantly influence the resulting mechanical properties and pore structure \cite{ZOLFAGHARI2018813, Bohr2023}. However, its effect on porosity may be more variable and cannot be reflected from the current compression data, which may exhibit large variance due to continuous system tuning. A more in-depth analysis will be presented in Section \ref{Effect of Humidity on Membrane Morphology and Mechanics}.

\subsection{Effect of Polymer Concentration on Membrane Morphology and Mechanics}

\begin{figure*}[!ht]
\centering
\includegraphics[width=\linewidth]{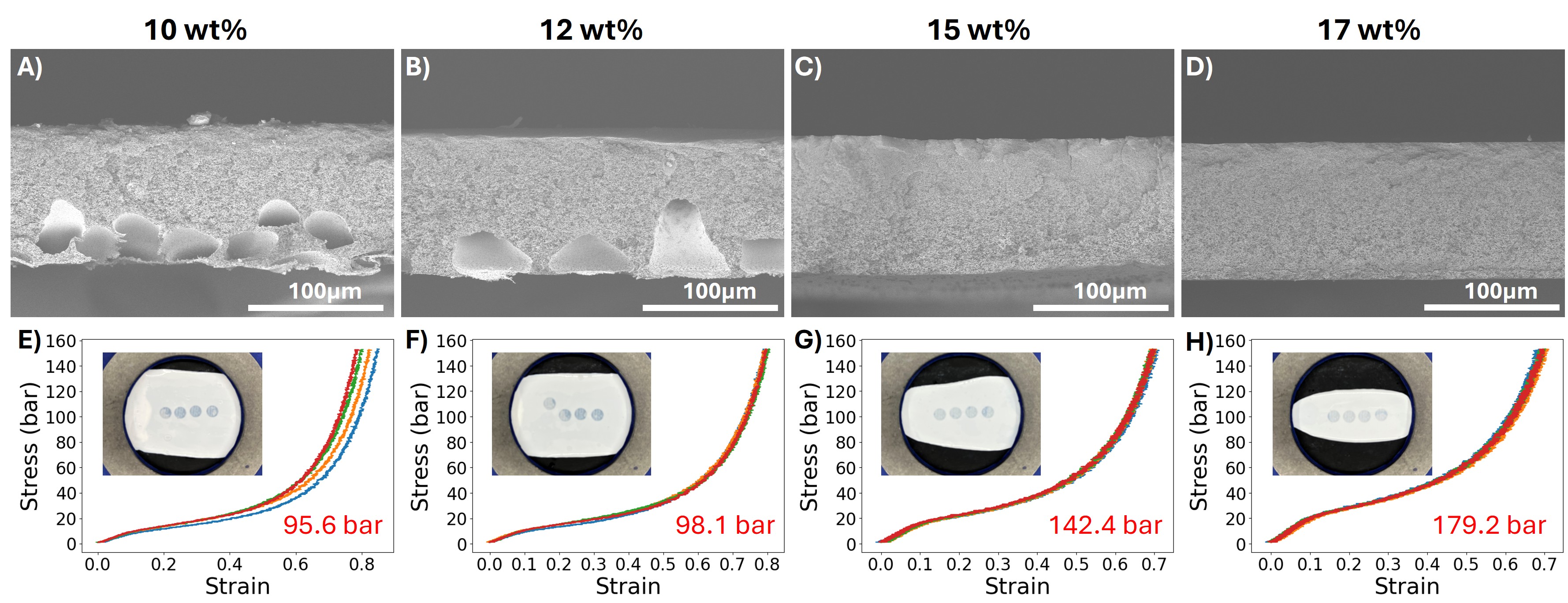}
\caption{SEM cross-sectional images (A--D) and corresponding stress-strain curves from multiple compression tests (E--H) of membranes fabricated at varying polymer concentrations: (A, E) 10~wt\%, (B, F) 12~wt\%, (C, G) 15~wt\%, and (D, H) 17~wt\% at a relative humidity of 64~RH\%. The extracted elastic modulus is annotated in each stress-strain plot. SEM images reveal clear structural transitions from highly porous membranes containing large macrovoids at lower concentrations (10~wt\% and 12~wt\%) to denser, sponge-like structures at higher concentrations (15~wt\% and 17~wt\%). The stress-strain curves (E--H) demonstrate increasingly uniform mechanical responses and enhanced mechanical stiffness as the polymer concentration increases.}
\label{concentration.jpg}
\end{figure*}

Given the observed trend of polymer concentration on the mechanical properties and pore structure of membranes, a detailed examination is warranted. Figure~\ref{concentration.jpg} presents the effect of polymer concentration on membrane morphology and mechanical performance. All samples were fabricated using the automated platform, including polymer solution preparation via the dilution-based approach. Cross-sectional SEM images show a clear morphological transition with increasing polymer concentration. At 10~wt\% and 12~wt\%, the membranes exhibit large macrovoids at the bottom, which are a characteristic outcome of rapid demixing during NIPS. As the concentration increases to 15~wt\% and 17~wt\%, the morphology becomes pure sponge-like as expected from the slower demixing.

Shown in Figure~\ref{concentration.jpg} (E--H), these morphological differences are reflected in the mechanical behavior of the membranes. Stiffness, as indicated by the slope of the elastic region, increases with polymer concentration. Compressibility, defined as the extent of deformation under load, decreases, which is consistent with reduced porosity at higher concentrations. The 10~wt\% membrane exhibits noticeable variability among the curves, with this intra-sample variability likely due to structural heterogeneity and the presence of large macrovoids. In contrast, membranes fabricated at higher polymer concentrations show more tightly aligned curves, indicating high intra-sample uniformity.

These results also reproduce the strong influence of polymer concentration on membrane morphology and mechanical behavior. Lower concentrations lead to rapid phase separation and highly porous morphologies that may enhance water permeability but introduce mechanical variability and weakness \cite{Tan2019}. Higher concentrations may prevent macrovoid formation and promote denser, more uniform structures with greater stiffness and lower compressibility, which are favorable for high-pressure applications, although potentially at the cost of permeability \cite{Tan2019}. The ability of the automated platform to reproduce these well-established structure-property relationships further demonstrates its effectiveness and reliability for systematic membrane development.

\subsection{Effect of Humidity on Membrane Morphology and Mechanics}
\label{Effect of Humidity on Membrane Morphology and Mechanics}
\begin{figure*}[!ht]
\centering
\includegraphics[width=\linewidth]{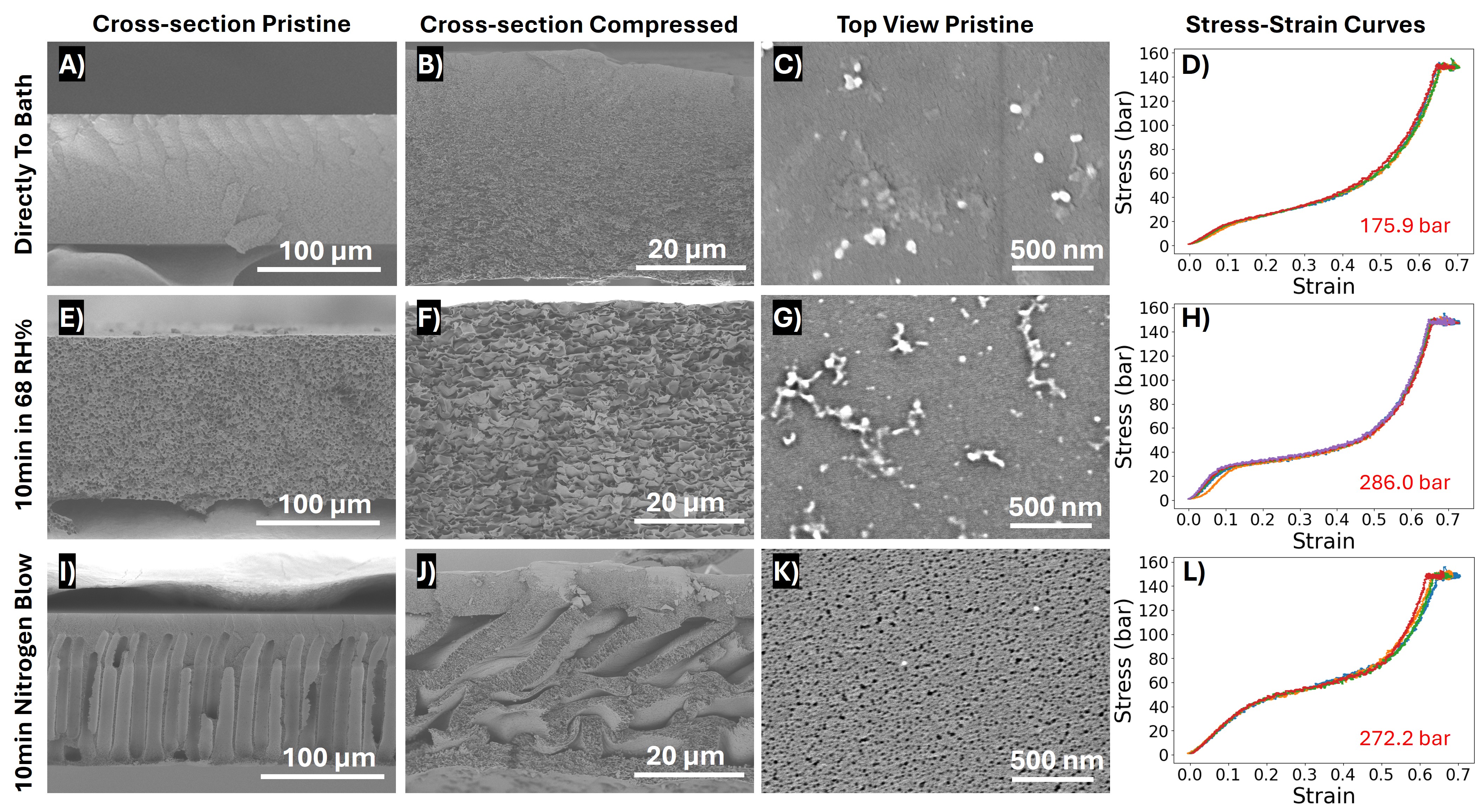}
\caption{SEM images of membrane cross-sections and top views and stress-strain curves under different fabrication conditions with 17~wt\% solution: (A--D) direct immersion into the coagulation bath, (E--H) exposure to 68~RH\% for 10~min before immersion, and (I--L) nitrogen blowing for 10~min before immersion. (A, E, I) Pristine cross-section images of membranes fabricated. (B, F, J) Cross-section images after compression testing to a maximum stress of 150~bar, (C, G, K) Top-view SEM images reveal surface morphology and pore structures. (D, H, L) stress-strain curves obtained from compression testing with extracted elastic moduli annotated.}
\label{sems.jpg}
\end{figure*}

To illustrate the effects of humidity during casting and to demonstrate the ability of the system to control the environmental humidity, Figure~\ref{sems.jpg} shows the membrane morphologies and the corresponding stress-strain curves under three distinct casting conditions: (1) immediate immersion into the coagulation bath, (2) exposure to ambient humidity at 68\%~RH for 10~min before immersion, and (3) exposure to a dry nitrogen flow for 10~min before immersion. 

For membranes directly immersed without intermediate treatments (Figure~\ref{sems.jpg}A--D), the pristine cross-section shows a dense sponge-like morphology without any observable macrovoids (Figure~\ref{sems.jpg}A). The absence of macrovoids is attributed to pre-existing moisture absorption within the polymer solution, resulting in slower and more uniform demixing upon immediate immersion \cite{Padilha2019}. After compression (Figure~\ref{sems.jpg}B), the sponge-like porous structure is uniformly compacted without noticeable structural collapse or localized deformation. The top-view SEM image (Figure~\ref{sems.jpg}C) shows a relatively smooth membrane surface but with no visible pores; a small amount of particulate debris is observed. Stress-strain curves (Figure~\ref{sems.jpg}D) show moderate stiffness (modulus 175.9~bar) with a gradually sloping plateau region, suggesting continuous pore collapse without distinct structural transitions.

Exposure to high ambient humidity (68~RH\%, Figure~\ref{sems.jpg}E--H) significantly alters membrane morphology by premature phase separation of the polymer solution prior to immersion, as evidenced by the polymer film turning opaque during the humidity exposure step. The pristine cross-section (Figure~\ref{sems.jpg}E) reveals a highly porous cellular structure. This cellular morphology arises from the premature moisture absorption in the polymer solution film, initiating a slow phase separation process \cite{Baldo2020}. Under compression (Figure~\ref{sems.jpg}F), the cellular pores deform uniformly without localized collapse. The top-view SEM image (Figure~\ref{sems.jpg}G) shows no visible surface pores; the white debris present is likely due to surface contamination. Stress-strain curves (Figure~\ref{sems.jpg}H) exhibit a stiffer elastic region (modulus 286.0~bar) followed by a distinctively flat plateau region, indicative of uniform pore collapse and low structural resistance of the large cellular pores.

Membranes subjected to dry nitrogen exposure (Figure~\ref{sems.jpg}I--L) exhibit the formation of finger-like macrovoids beneath a dense sponge-like skin layer (Figure~\ref{sems.jpg}I). Dry nitrogen treatment effectively removes residual moisture from the polymer film, promoting faster solvent evaporation and facilitating rapid solvent-nonsolvent exchange upon immersion \cite{Tan2019, DEHBAN202254}. Post-compression cross-sectional imaging (Figure~\ref{sems.jpg}J) reveals that structural deformation predominantly occurs within these macrovoid regions, which collapse adjacently and stack under compressive load. This pore collapse mechanism may create localized resistance, redistributing the stress and concentrating it at curved or folded pore boundaries. Such stress concentrations can likely enhance local stiffness (modulus 272.2~bar) and overall mechanical strength, as suggested by the corresponding stress-strain curves (Figure~\ref{sems.jpg}L). This observation is contradictory to the established notion that macrovoids typically degrade mechanical performance under high-pressure conditions \cite{Yip2010}. The top-view SEM image (Figure~\ref{sems.jpg}K) reveals a highly uniform porous membrane surface, indicating that the reduced humidity from dry nitrogen treatment promotes desirable surface porosity, which is beneficial for enhancing membrane permeability in filtration applications. 

These results highlight the critical impact that humidity has on the membrane morphology and corresponding mechanical properties during NIPS fabrication. Immediate immersion without intermediate treatments results in dense, sponge-like membranes with generally high stiffness but limited surface porosity. Exposure to elevated humidity promotes premature phase separation, forming large cellular pores but reducing mechanical stiffness. In contrast, pre-treatment with dry nitrogen rapidly removes existing moisture in the film, leading to the formation of finger-like macrovoids under a sponge-like skin layer, unexpectedly improving mechanical stiffness despite conventional expectations of macrovoid-induced structural weaknesses. Whether this contributes positively or negatively to long-term performance in filtration applications remains uncertain, but it presents an intriguing structural feature that is worth further investigation. Understanding how these localized deformation mechanisms influence membrane durability under operational stress could inform new design strategies that leverage pore geometry for performance optimization. 

\subsection{Surface Pore Size Distribution of Nitrogen-Treated Membranes}
\label{PSD}
\begin{figure}[!ht]
\centering
\includegraphics[width=\linewidth]{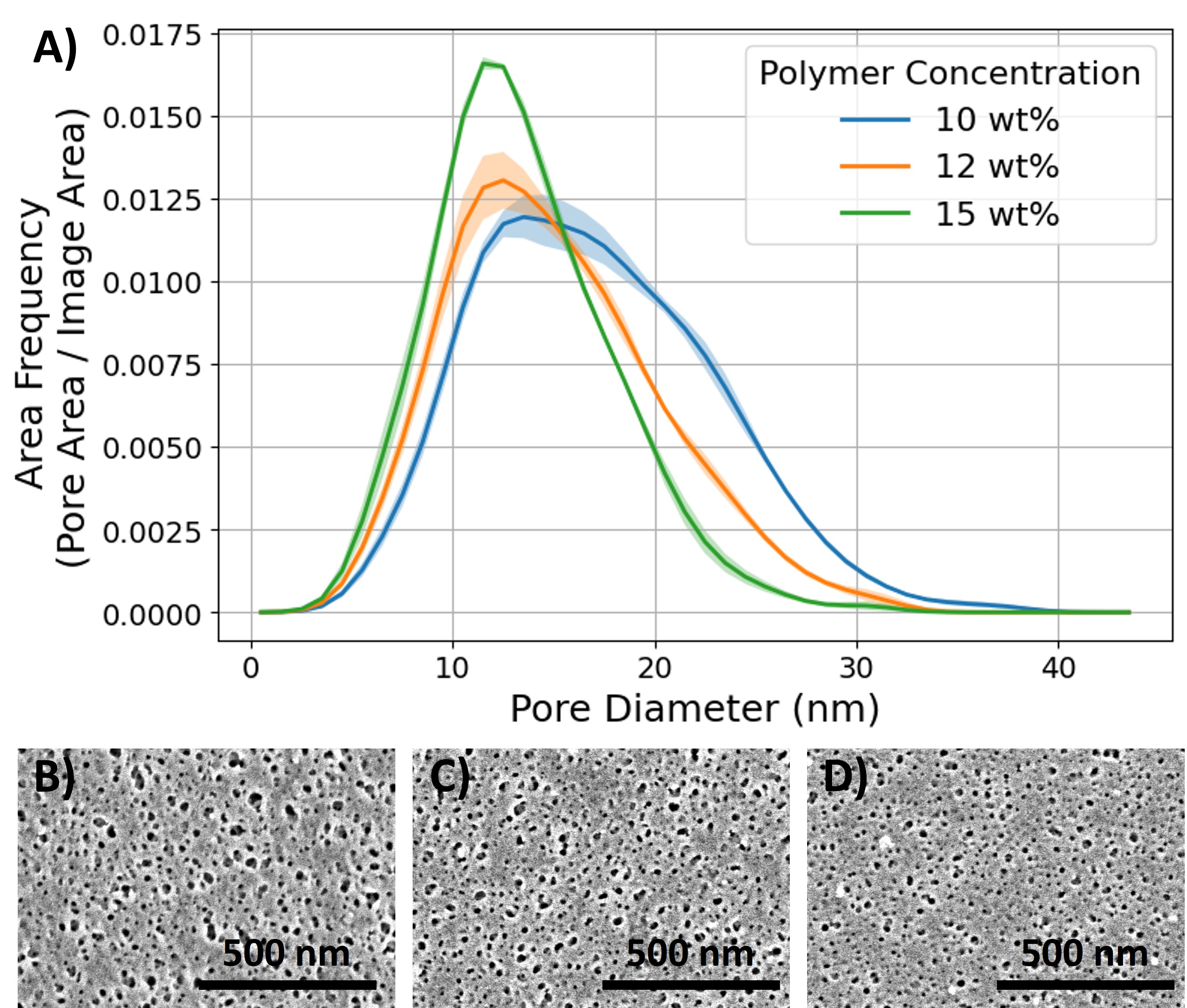}
\caption{Surface PSD analysis of nitrogen-treated PSf membranes. A) Surface PSD of PSf membranes fabricated at 10, 12, and 15~wt\% polymer concentrations under 10 min dry nitrogen blow sing the automated platform. Each condition includes three measurements taken from different regions. The plot approximates the uncoated pore sizes by correcting for sputter‑coating‑induced pore narrowing \cite{SIMA2025}. The solid lines represent the mean PSD for each condition and the shaded regions indicate the standard deviations. Surface SEM images of membranes fabricated at B) 10~wt\%, C) 12~wt\%, and D)15~wt\% polymer concentrations. Binary images of the surface pore structures are presented in the Supplementary Information.}
\label{PSD.jpg}
\end{figure}

Knowing that nitrogen-treated membranes yield desirable surface pore structures, we further quantified how polymer concentration affects these features by conducting a pore size distribution (PSD) analysis. Nitrogen-treated PSf membranes were synthesized using the fully automated platform, which integrates automated dilution and casting steps to ensure precise and consistent sample preparation. For each concentration (10~wt\%, 12~wt\%, and 15~wt\%), PSD measurements were taken at three different surface locations to capture spatial variability. As shown in Figure~\ref{PSD.jpg}, the PSD curves consistently shift toward larger diameters and broaden with decreasing polymer concentration. 

This indicates that membranes made from more dilute polymer solutions tend to exhibit larger and more heterogeneous surface pores, consistent with prior observations in the literature \cite{ZOLFAGHARI2018813}. In addition to PSD, surface porosity was also quantified from the thresholded SEM images (Table \ref{tab:surface_porosity}), showing a systematic decrease with increasing polymer concentration. The consistency of the observed trends across different regions of each membrane highlights the spatial uniformity of the pore formation process under controlled nitrogen exposure and fabrication processes, further demonstrating the reliability and repeatability of the automated platform in maintaining controlled fabrication conditions. 

\begin{table}[htbp]
\centering
\caption{Surface porosity from SEM images (mean $\pm$ standard error; $n=3$). Uncoated values corrected for 1.8~nm sputter coating thickness.}
\label{tab:surface_porosity}
\begin{tabular}{lcc}
\hline
Polymer conc. & Coated & Uncoated \\
\hline
10~wt\% & 11.0 $\pm$ 0.2 & 18.7 $\pm$ 0.5 \\
12~wt\% &  9.2 $\pm$ 0.8 & 16.9 $\pm$ 1.5 \\
15~wt\% &  8.7 $\pm$ 0.3 & 17.0 $\pm$ 0.5 \\
\hline
\end{tabular}
\end{table}

\FloatBarrier
\section{Conclusion} 

This work presents the development and validation of a high-throughput, fully automated NIPS platform that enables reproducible fabrication and mechanical characterization of polymer membranes. By integrating liquid handling, controlled casting, environmental modulation, and compression testing, the system streamlines membrane synthesis from solution preparation to rapid mechanical screening with minimal human intervention. The modular architecture of the platform allows for precise control over critical fabrication parameters such as polymer concentration and ambient humidity, and supports parallel processing to further increase experimental throughput. Compression testing is used as a practical and sensitive proxy for assessing internal structure and mechanical consistency. The automatic segmentation of stress-strain curves enables extraction of key descriptors such as elastic modulus and pore fraction, which correlate with observed structural features and porosity trends. Although this method does not directly assess the filtration performance, it provides high-resolution mechanical data that are indicative of membrane uniformity and pore structure. Key findings from this study reproduce established works in which polymer concentration and humidity strongly influence membrane morphology and mechanical properties. Higher polymer concentrations lead to denser sponge-like structures with increased stiffness and uniformity, while humidity impacts pore structure by altering phase separation kinetics. Notably, membranes exposed to dry nitrogen during casting exhibited finger-like macrovoids yet surprisingly high stiffness and strength, suggesting complex micro-mechanical interactions or localized densification during compression. These results highlight the importance of controlling both the formulation and environmental factors to achieve the targeted membrane performance. The platform's capability for closed-loop experimentation lays the foundation for accelerated membrane development. Its compatibility with algorithmic optimization workflows, including machine learning and active learning strategies, positions it as a valuable tool for exploring complex fabrication spaces. Future work should focus on exploring more fabrication parameters and integrating real-time predictive models to enable fully autonomous discovery and optimization of membrane formulations.

\section*{CRediT authorship contribution statement}
\textbf{Hongchen Wang:} Writing - Original Draft, Writing - Review \& Editing, Methodology, Investigation, Formal analysis, Data curation, Conceptualization, Visualization, Validation, Software. \textbf{Sima Zeinali Danalou:} Writing - Review \& Editing, Methodology, Software, Formal analysis, Data Curation, Visualization. \textbf{Jiahao Zhu:} Methodology, Visualization. \textbf{Kenneth Sulimro:} Methodology, Software. \textbf{Chaewon Lim:} Methodology, Software. \textbf{Smita Basak:} Methodology, Software. \textbf{Aim\'ee Tai:} Methodology, Software. \textbf{Usan Siriwardana:} Methodology, Software. \textbf{Jason Hattrick-Simpers:} Writing - Review \& Editing, Project administration, Funding acquisition, Supervision, Resources, Conceptualization. \textbf{Jay Werber:} Writing - Review \& Editing, Project administration, Funding acquisition, Supervision, Resources, Conceptualization.

\section*{Declaration of competing interest}
The authors declare no conflict of interest.

\section*{Declaration of generative AI and AI-assisted technologies in the writing process}
During the preparation of this work, the authors used ChatGPT (OpenAI) in order to assist with language refinement, rewording, and improving readability. After using this tool, the authors reviewed and edited the content as needed and take full responsibility for the content of the published article.

\section*{Acknowledgment}
This work is financially supported by NSERC through the Alliance Missions program. The authors thank Steve Harrold and Andrew Back of Veolia Water Technology \& Solutions for valuable technical feedback. The authors acknowledge the Open Centre for the Characterization of Advanced Materials (OCCAM) at the University of Toronto for providing instrumental support for SEM characterization in this study. The authors thank William Luecke for helping interpret the compression test results and for revising the manuscript. The authors thank Yingxi Wang for refining the graphical abstract.

\appendix
\section{Supplementary data}
Supplementary data to this article can be found online at (to be inserted).

\section*{Data availability}
Data will be made available on request.
{\small
\bibliographystyle{elsarticle-num}
\bibliography{main.bib}
}

\end{document}